\documentclass[smallextended,final]{svjour3}
\usepackage[utf8]{inputenc}
\usepackage[greek,english]{babel}
\usepackage{graphicx}
\usepackage{textcomp}
\usepackage{color}
\usepackage{amssymb,amsmath}
\usepackage{hyperref}
\usepackage{times}
\usepackage[authoryear,round]{natbib}
\usepackage{multirow}
\usepackage{arydshln}
\usepackage{enumitem}
\newcommand{\comment}[1]{}

\title{Reliability and Interpretability in Science and Deep Learning}
\author{Luigi Scorzato}
\institute{Accenture AG, Gen{\`e}ve, Switzerland.
\email{luigi.scorzato@accenture.com}}
\date{}

\begin{document}

\newtheorem{defn}{Definition}
\newtheorem{ass}{Assumption}
\newtheorem{rmrk}{Remark}

\maketitle

\begin{abstract}
In recent years, the question of the reliability of Machine Learning (ML) methods has acquired
significant importance, and the analysis of the associated uncertainties has motivated a growing
amount of research.  However, most of these studies have applied standard error analysis to ML
models---and in particular Deep Neural Network (DNN) models---which represent a rather significant
departure from standard scientific modelling.  It is therefore necessary to integrate the standard
error analysis with a deeper epistemological analysis of the possible differences between DNN
models and standard scientific modelling and the possible implications of these differences in the
assessment of reliability.  This article offers several contributions.  First, it emphasises the
ubiquitous role of model assumptions (both in ML and traditional science) against the illusion of
theory-free science.  Secondly, model assumptions are analysed from the point of view of their
(epistemic) complexity, which is shown to be language-independent.  It is argued that the high
epistemic complexity of DNN models hinders the estimate of their reliability and also their
prospect of long term progress.  Some potential ways forward are suggested.  Thirdly, this article
identifies the close relation between a model's epistemic complexity and its interpretability, as
introduced in the context of responsible AI.  This clarifies in which sense---and to what
extent---the lack of understanding of a model (black-box problem) impacts its interpretability in
a way that is independent of individual skills.  It also clarifies how interpretability is a
precondition for a plausible assessment of the reliability of any model, which cannot be based on
statistical analysis alone.  This article focuses on the comparison between traditional scientific
models and DNN models.  However, Random Forest (RF) and Logistic Regression (LR) models are also
briefly considered.
\end{abstract}

\section{Introduction}
Machine Learning (ML) methods in general \citep{hastie_09_elements} and Deep Neural Networks
(DNNs) in particular \citep{Goodfellow-et-al-2016, lecun2015deeplearning} have achieved tremendous
successes in the past decade.  For example, a classifier based on the {\em resnet} architecture
\citep{resnet2015} reached human level accuracy in the ILSVRC2015 challenge \citep{ILSVRC2015}.
Furthermore, Neural Networks based on the idea of {\em transformers} \citep{Vaswani2017attention}
have recently spurred breakthroughs in the field of Natural Language Processing (NLP), enabling
high-quality machine translation.  Answers generated by Large Language Models (LLM) such as GPT-3
\citep{openai-gpt3} have achieved an impressive level of similarity to those generated by humans.
There is by now convincing evidence that the best ML/DNN algorithms can learn effectively highly
sophisticated tasks.  However, some important questions remain open concerning the reliability of
DNN algorithms.

First, we cannot completely exclude the possibility that successful DNN algorithms are
over-fitting the collections of datasets that are used to train and test them
\citep{Berkeley2018overfit}.  In fact, because of the difficulty of collecting good quality
labelled data, very few very popular datasets (in particular CIFAR-10, \citet{cifar10}, ImageNet,
\citet{ILSVRC2015} and a few others) represent the unique benchmark for the majority of the
research work on DNN.  This challenges one of the key statistical assumptions of all ML methods,
namely that the parameters are set independently of the test data.  In principle, {\em nothing}
about the test data should be known in the design and training phase.  In practice, train-test
contamination can occur in subtle ways, even if we don't directly use test data for training
\citep{kapoor_leakage_2023}.  Strictly speaking, if some data have been used in an article that we
read before designing our new DNN architecture, those data should not be used to test our new
architecture, because they might already have influenced our design choices indirectly.  In
practice, it can be hard to adhere to this requirement rigorously.

Secondly, successful applications of ML methods are much more likely to be published than
unsuccessful ones.  While {\em publication biases} \citep{pubbias1987} affect all types of
scientific research, one might expect a stronger impact on ML research, because ML models are less
likely to bring valuable insight when they fail.  In fact, we have less compelling reasons to
believe that they {\em must} succeed, which is related to the lack of full understanding of the
mechanism that allows a DNN model to learn some features and ignore others.

Furthermore, assessment of confidence levels of ML predictions is notoriously difficult,
especially in the case of DNN.  The best evidence of these difficulties is provided by so-called
{\em adversarial examples} \citep{szegedy2014intriguing}: images that are misclassified with very
high confidence by the DNN classifier, although they are humanly indistinguishable from other
images that are correctly classified.  Adversarial examples are much more pervasive in DNN than
initially expected \citep{carlini2017adver}.  The problem is not the occurrence of
misclassification itself, but the very high confidence (easily $>99\%$) quoted by the classifier.
This is clearly not a reliable error estimate\footnote{It also does not help to say that
adversarial examples have negligible probability among natural images, because such statement
cannot be made quantitative, as we lack a suitable characterisation of the space of {\em possible
  natural images}.  Moreover, adversarials have been also reproduced from natural images
\citep{Hendrycks2019NaturalAdversarial}.}.

Finally, the presence of social biases in some datasets widely used to train ML algorithms
\citep{gendershades} is a matter of concern.  Reliable error estimates could help significantly to
detect earlier those predictions that rely on too limited statistics.  This would be an important
step toward the goal of deploying responsible AI \citep{EUAIact2023, ec2019ethics} at scale.

ML methods in general, and DNN in particular, are being used successfully in many contexts where
assessing their expected errors is not necessary.  This applies, for example, to all those cases
where the ML method improves the search efficiency of candidate solutions that can be subsequently
verified by other means \citep{duede_2023}.  In these cases, ML methods are used in the context of
discovery, and not in the context of justification.  Examples cover a vast range of applications,
spanning material discovery, drug discovery, predictive maintenance, fraud detection
\citep{baesens2015fraud}, code suggestion \citep{openai-gpt3}, protein folding \citep{AlphaFold}
and many others.  In those cases, assessing the reliability is the responsibility of the
independent check, while the ML method merely improves the efficiency of the overall process.
However, there are also applications where independent checks are not practical (e.g. for
safety-critical real-time systems \citep{buttazzo2022embedded}) or cases where it is crucial to
estimate the amount of missed solutions (e.g. compliance applications).  In those cases, assessing
the reliability of the ML method is very important.

This paper addresses the reliability of DNN methods from a fundamental epistemological point of
view.  It is very important to combine this perspective with a purely statistical one because even
traditional science does not offer an absolute (assumption-free) guarantee that its predictions
have specific probabilities.  So, we must understand to what extent DNN models rest on similar
grounds as those employed by more traditional scientific disciplines and to what extent they might
suffer from fundamentally different problems.

This paper focuses on models based on DNNs, because these have been responsible for the most
impressive successes in recent years and because they pose the most interesting challenges from an
epistemological point of view.  However, we will occasionally consider also other ML methods, such
as Logistic Regression (LR) and Random Forest (RF) models as they share some features of DNN
models and some features of traditional scientific models.  For the sake of concreteness, we focus
on supervised ML models for binary classification.  However, everything discussed in this article
could be easily extended to other models of supervised ML.  Unsupervised ML methods have quite
different purposes and are beyond the scope of this paper.

The topic of this article falls at the intersection of the theory of complexity
\citep{franklin2020algorithmic, Zenil2020review, LiVitanyi2019, Hutter2007ScholarpediaAIT}, DNNs
\citep{Goodfellow-et-al-2016, lecun2015deeplearning}, responsible AI \citep{AIandEthics,
  eitel2021beyond} and the epistemology of ML, which has attracted considerable attention recently
(see e.g.~\citet{Floridi2022-FLOTEF, sep-science-big-data}).  As opposed to many works on this
topic (but similar to \citet{SyntheseCollPoS-AI, zenil2020epistDL, grote2024reliability} among others), this article
stresses the important peculiarities of DNNs within ML methods.  The present approach has some
similarities with the one adopted in \citep{watson2021explanation}.  Some of the most important
differences are: (a) the present focus is on {\em reliability}, that naturally leads to consider
{\em global}, rather than {\em local} interpretability; (b) for the same reason, the {\em
  relevance} defined in \citep{watson2021explanation} is less applicable and it is not considered
here; (c) finally, \citet{watson2021explanation} consider a subjective notion of simplicity, while
we focus on the hardcore complexity that no human can reduce, regardless of language and
individual skills.  This article investigates the objective foundations of reliability assessment
(i.e. based on well-defined assumptions).  Hence, it is largely unrelated to the literature that
investigates the sociological basis for the trust in ML models, or analogies between ML and human
behaviours (see e.g.,~\citet{duede2022instruments, clark2022decentring, tamir2023machine}).
Further comparisons with the literature are provided in the main text.

\section{Assessing the reliability of model predictions in science and ML}
\label{sec:compare}

To assess the reliability of any model we must be able to estimate the uncertainty of its
predictions, in some precise and useful probabilistic sense.  To start, it is convenient to
distinguish between {\em statistical} and {\em systematic} uncertainties.  Traditionally,
statistical uncertainties are defined as those error sources which have a known statistical
distribution \citep{bohm2017introduction}\footnote{The ML literature adopts sometimes similar but
slightly different classification between statistical (aka random, aleatoric or `data')
uncertainties on one side and systematic (aka epistemic or model) uncertainties on the other
\citep{hullermeier2021aleatoric, IEEE_errors_review}.  Here we stick to the traditional one,
because it clarifies the role of the model assumptions, which is crucial in our discussion.}.
These uncertainties can be safely analysed via statistical methods.  Systematic uncertainties are
all the others: they may stem from systematic distortions in the measurement devices, from
imbalances in the data selection, from inaccurate models, from approximation errors, or from
inaccurate parameter fitting.  As emphasised in \citet{bohm2017introduction}, even random noise
with unknown statistical properties must be classified as a systematic effect\footnote{Note that
probability distributions without finite mean and variance, for which the central limit theorem
does not apply, cannot be characterised empirically in most cases.  Given the ubiquity of, e.g.,
self-similar phenomena \citep{barenblatt1996scaling} it is clear that one cannot certainly assume
that every noise generates {\em statistical} uncertainties.}.  Statistical uncertainties can be
very difficult to estimate in practice, but the process is conceptually clear.  They are {\em
  known-unknown}, as they are expected by the model.  Systematic uncertainties are {\em
  unknown-unknown} and they can be estimated only by enriching the model assumptions.  This raises
the question of {\bf which assumptions are acceptable to assess our model assumptions?}

This question is crucial and problematic not only for ML models but also for traditional
scientific (TS) models\footnote{Let us be more specific on what we mean by TS models.  These
include any state-of-the-art textbook theories such as, e.g., the standard model of particle
physics.  Because we focus here on DNN models for classification, it may be useful to keep in mind
also examples of TS models of classifications, such as the model of stellar classification
\citep{gray2009stellar} or the classification of chemical elements \citep{Emsley} or the medical
classification of diseases \citep{ICD11}.  In general, TS models are based on some well defined
features, whose relevance is based of a wealth of background science.}, and they have been the
focus of extended analysis in the philosophy of science.  It is therefore important to compare ML
models to TS models in this respect, to understand to what extent ML introduces novel issues to
the existing ones.  It turns out that studying the reliability of ML models forces us to
reconsider classic philosophical questions that are too often regarded as void of practical
consequences.

Besides DNN and TS models, in the present comparison, we consider also Logistic Regression (LR)
models and Random Forest (RF) models, because they represent very popular ML models whose
properties often interpolate between DNN models and TS models in an interesting and instructive
way.

\subsection{Sources of errors}
\label{sec:sources}

It is useful to distinguish four different sources of errors and see how they affect ML and TS
models in different ways.  Errors may stem (a) from data measurements, including the labels of
training data; (b) from the model, which may not faithfully describe the actual phenomena in
scope; (c) from any approximation that may be applied to the model to derive further predictions;
and (d) from fitting the parameters that are left free by the model\footnote{Type (a) and (d) may
have both {\em statistical} and {\em systematic} contributions, while type (b) and (c) are purely
{\em systematic}.}\footnote{Models with free parameters are, strictly speaking, not yet specific
enough to enable the derivation of predictions.  They can be considered model-classes, but we
simply use the term ``model'' so as not to burden the notation.}.

Source (a) is the same for any models under consideration (ML or TS).  Note that all measurements
are theory-laden \citep{hanson1965patterns, Duhem1954}.  But, whenever two models are employed to
study the same phenomena, we can also assume that the theory behind these measurements is the
same.  Source (c) applies only when the original model is replaced by an approximation to enable
further analytic or numerical derivation.  ML models are already in a form that can be used for
numerical applications and therefore source (c) applies only to some TS models.

Sources (b) and (d) are the most interesting ones when comparing TS and ML models.  Model errors
(b) can be reduced by extending the basic (ideal) model with further elements designed to model
the deviations from the basic model.  However, this enrichment comes at the expense of
more complex assumptions and more parameters to be fitted (generating more uncertainties of type (d)).

Source (d) refers to errors in the determination of the optimal parameters, even when the model
might provide a correct representation of the phenomena.  This may happen because of limited data,
noisy data (source (a)), an inaccurate minimisation algorithm or an inexact parameter
fitting process.  Error source (d) has a statistical component, which is propagated from noisy
data (a) and can be analysed statistically (along the lines of e.g.~\citet{MarshallBlencoe2005}
and references therein).  However, source (d) also has systematic components, due to the
complexity of determining the optimal parameters and also due to the unknown statistical
properties of the data\footnote{for some classes of noise sources, collecting sufficient data can
be completely hopeless in practice (see, e.g., \citet{deForcrand:2009zkb}) or even in principle
(see e.g., \citet{gabaix2009power}).}.  Moreover, increasing the number of parameters requires, in
general, an exponential increase in the amount of data required for their
determination\footnote{An exponential increase is expected, in the generic case, according to the
curse of dimensionality \citep{giraud2021highdimstat}.}.  In conclusion, also estimating (d)
requires making suitable assumptions.

TS and ML models do not show fundamental {\em qualitative} differences from the point of view of
the error sources above (except for the less important source (c)).  But they do display {\em
  quantitative} differences: ML models tend to be very flexible and characterised by significantly
more parameters than TS models.  Traditionally, fitting too many parameters was considered
hopeless, because of the curse of dimensionality \citep{giraud2021highdimstat}.  But, the
impressive predictive success of DNN models, despite having vastly more parameters than training
data points, shows that we cannot dismiss them based on naive parameter counting.  The question of
this paper is whether we can also ensure the reliability of models with so many parameters.  We
will see that this is a very different question that will be addressed in Sec.~\ref{sec:DNNerrors}
and Sec.~\ref{sec:ass-int}.  But, before coming to this, we comment, in Sec.~\ref{sec:ill}, on the
radical idea that the success of DNNs might even represent a new kind of ``theory-free'' science.

\subsection{The illusion of predictions without assumptions}
\label{sec:ill}

A common fallacy is to hope that we can produce error estimates which do not depend on any
assumption, or, equivalently, produce estimates that take into account all possible models.  In
the context of ML, this fallacy becomes even more tempting because ML models can be very flexible,
and, for some DNN models, a universal approximation theorem applies \citep{univ-approx-th,
  leshno1993univ-approx}.  The idea of ``theory-free'' science does enjoy some support
\citep{anderson2008theoryfree} and it is not ruled out by others \citep{Floridi2022-FLOTEF}.

The simple reason why this idea is fallacious is the well-known (but often overlooked) {\em
  underdetermination of the theory} by the data \citep{Duhem1954, quine1975empirically,
  sep-scientific-underdetermination}.  No matter how much data we have collected, there are always
infinite different models that fit those data within any given approximation.  Moreover, for every
conceivable prediction, there exist infinite models that still fit all the previous data and also
deliver the desired prediction.  Data {\em alone} cannot enforce {\em any} conclusion.

Even ML models that enjoy the universal approximation property offer no exception to the
underdetermination thesis\footnote{Even assuming that the training process is able to attain the
best approximation, which is not ensured by the universal approximation theorem itself.}.  On the
contrary, they provide more evidence for it, because we can include any desired prediction among
the data that we want to describe and the universal approximation property will ensure the
existence of parameters that reproduce also the desired prediction within the desired
approximation.  The predictions of the ML model hence depend not only on the data but also on the
architecture of the ML model and, in the absence of evidence to the contrary, also on the
algorithmic and initialisation details.

Another frequent fallacy is the idea that we should only accept assumptions that have been tested.
But how much and under which conditions should they be tested? Newton's laws were tested
extensively under all conceivable circumstances for over two centuries before discovering that
they do not always hold.  There is no way to fully test our assumptions (or to define their scope
of applicability) in a future-proof way.  They may fail in contexts that we cannot currently
foresee.

In conclusion, any error estimate is necessarily based on assumptions that cannot be fully
justified on an empirical basis.  They might be narrower or broader than the assumptions of the
model itself, but they must be acknowledged because the error estimates crucially depend on them.

\section{Current approaches to assess the reliability of DNN predictions}
\label{sec:DNNerrors}

After the general considerations of the previous section, let us discuss how the reliability of
DNN predictions is currently assessed in the literature.

Until recently, it was very common to quote the normalised exponential function (or softmax)
\citep{Bishop2006} as a measure of confidence in a DNN prediction.  However, it has been shown
very convincingly that the softmax output lacks the basic features of a measure of confidence (see
e.g. \citet{Guo2017}). In particular, it tends to be increasingly overconfident for increasingly
out-of-distribution samples \citep{hein2019relu}.  Ad-hoc attempts to correct this pathology lead
to arbitrary results.

The issue of proper estimation has received growing attention in recent years, and numerous review
articles are available today \citep{IEEE_errors_review, IF2021review, loftus2022uncertainty}.  A
popular view regards the Bayesian approach as the most appropriate framework, in principle
\citep{wang2016bayesian}.  According to this view, the main drawback of a full Bayesian estimate
is its prohibitive cost, which leads to a very active search for approximations that offer the
best trade-off between accuracy and computational efficiency \citep{titterington2004bayesian,
  BNN2020, MacKey1992bnn, gal2016dropout, blundell2015weight, sensoy2018evidential,
  hartmann2023transgressing}.  Besides the Bayesian framework, the other main approaches rely
either on ensemble methods \citep{Blundell2017ensamble, Wen2020BatchEnsemble, NN-bootstrap,
  Tavazza2021incpred}, or data augmentation methods \citep{shorten2019survey_DataAg,
  Wen_2021_DataAg}.  An alternative is to train the DNN to specifically identify outliers or
uncertain predictions (see e.g.~\citet{malinin2018predictive, raghu2019direct}).  Let us consider
them separately in the following.

\subsection{Bayesian Error Estimates}
\label{sec:BEE}

Consider the Bayesian approach first.  The computational cost is not the only difficulty.  It is
well known that Bayesian error estimates depend on the assumption of {\em prior distributions} (or
priors) for the parameters whose impact should be estimated, and the choice of priors is largely
arbitrary \citep{gelman2013bayesian}.  For infinite statistics, the Doob and Berstein - von Mises
theorems \citep{van2000asymptotic} state that the posterior distributions converge to
prior-independent limits.  In simple words: ``the data overwhelms the priors''.

However, such an argument is based on the unrealistic idea that the model is fixed, while we can
rather easily collect more data.  This is not what usually happens in real applications.  In many
cases, instead, collecting valuable data can be very hard, while producing new models is often
much easier. This is the case in many TS domains, where planning a new experiment might take
decades, while new models are published every day.  It is the case also in most ML applications,
where collecting {\em labelled} data can be extremely labour intensive, while new ML models are
produced at the rate of millions per second by the ML algorithm.  Under these conditions, there is
no guarantee that the dependency on the priors will be sufficiently suppressed by the time the
model is used.

Increasing the number of parameters might look like a way to eliminate the need of changing the
model.  But, the Doob and Berstein - von Mises theorems do not apply when the number of parameters
is of the same order as the size of the dataset or more \citep{johnstone2010high,
  sims2010understanding}.  Extensions exist for high-dimensional problems
\citep{banerjee2021bayesian} and even infinite-dimensional (aka nonparametric) ones
\citep{ghosal2017fundamentals}.  However, those extensions rely on the assumption that most
parameters are very close to zero, which is achieved via LASSO priors or equivalent suppression
mechanisms \citep{banerjee2021bayesian}.  Note that this is a much stronger assumption than
requiring that all the relevant parameters live in {\em some} low-dimensional subspace of a
high-dimensional model.  A parameter suppression mechanism imposes a sparse realisation of the
{\em given} parametrisation.  As a result, even in the asymptotic limit of infinite statistics,
the posteriors depend on the choice of the parametrisation, similar to the case of low-dimensional
models.  In other words, nonparametric models are not a way to eliminate model dependency.

The limitations of nonparametric models confirm the conclusions of Sec.~\ref{sec:ill} about the
impossibility of estimating errors that take into account {\em all possible models}.  The Bayesian
approach is an excellent framework to discuss the probability of a model within a given class (say
$C$) of models, but it cannot justify the selection of one class $C$ of models over other options.

In principle, the class $C$ should include a model that describes well all past and future data
(what is sometimes called a {\em true model}\footnote{Note that a {\em true model} defined in this
way might or might not exist or might not be unique in the given parametrization.})\footnote{To be
precise, if the goal is to assess the reliability of a {\em model prediction}, then the class $C$
should contain a true model, as stated in the main text. If the goal is to {\em test a model},
then the requirement applies to the background models to be used in combination with the model to
be tested (i.e. the background part should include a true model).  See Sec.~\ref{sec:ass} for more
details on background models.}.  However, we cannot know which are the true models, of course, and
we have to resort to other criteria.

The ambiguity in the selection of the class $C$ applies to both TS and DNN models.  However, in
practice, the scientists tend to agree on which other models should be included to assess the
errors of a given TS model.  Ideally, they should include all the models that might provide
legitimate alternative descriptions of the relevant phenomena, because they offer different
compromises between accuracy (for different phenomena) and non-empirical epistemic virtues.  These
are sometimes referred to as state-of-the-art models, but the class is not well defined unless we
identify the appropriate non-empirical epistemic virtues.  We will come back to this point in
Sec.~\ref{sec:simp}, where we consider a definition of state-of-the-art based on the simplicity of
the assumptions (see also App.~\ref{sec:progress}).  We will also see that identifying a similar
class for DNN models is significantly more challenging.

\subsection{Frequentist Error Estimates}
\label{sec:FEE}

The frequentist approach \citep{Blundell2017ensamble, Wen2020BatchEnsemble, NN-bootstrap} is not
better equipped to answer the questions above \citep{sims2010understanding}.  In fact, the
frequentist error assessment relies on the same assumptions that characterise the model that has
produced the predictions (and, possibly, additional assumptions).  By construction, the
frequentist approach is not designed to discuss the probability of the selected model.
Frequentist error estimates do have a value, but only under the assumption that the model is valid
for some parameters in the neighbourhood of the value selected by maximal likelihood.

This limitation is especially problematic for a model that displays high sensitivity to details,
which are not understood.  Because any little change to the parameters, as it happens, e.g., for
any new step of training, may bring the system to a configuration for which the previous error
analysis is not relevant anymore.

\subsection{DNN-based Error Estimates}
\label{sec:DEE}

The spectacular and inexplicable success of DNNs represents a strong temptation to try to use DNN
models to answer just about any difficult question, provided that we have at least some labelled
examples to train the model.  The fact that we do not understand why DNNs work, or when they do,
makes almost any new application an interesting experiment in itself.

In this sense, using {\em another DNN} to detect outliers \citep{malinin2018predictive} or
uncertain predictions \citep{raghu2019direct} represents an interesting further application of
DNN, but it does not provide an error estimate that relies less on DNN model assumptions.  On the
contrary, we must rely on {\em two} DNN models, in this case.

On the other hand, the same criticism may be directed equally well against traditional science: we
rely on the laws of physics to build experimental devices that we use to test the laws of physics
themselves.  And we rely on other laws of physics to determine the accuracy of our measurements.
Why is this problematic for DNN, if it is not for traditional science?  Or should it be?  Again,
the analysis of DNN forces a reappraisal of classic epistemological questions concerning
traditional science.  We do it on Sec.~\ref{sec:ass-int}, but, before that, we consider one last
potential approach.

\subsection{Predictions are NOT all you need}
\label{sec:pred}

Can't we just use the success rate of the test dataset as an estimate of the expected prediction
error?  Intuitively, one expects that {\em novel predictions} on unseen data do provide a
meaningful test.  But, from a statistical point of view, the success rate corresponds to the
simplest frequentist statistical estimator, with the limitations already discussed in
Sec.~\ref{sec:FEE}.  From a philosophical point of view, the idea that successful tests confirm a
model is fraught with riddles \citep{sep-confirmation, confirmation-IEP}.  In fact, for example,
we cannot be sure that the future data will conform to the same distribution of the past data.
One might argue that this corresponds to a change of domain of application and the model cannot be
blamed for that.  But we cannot define the scope of applicability of any model in a way that is
future-proof: the model may fail in circumstances that we currently cannot even imagine.  All the
attempts to quantify the degree to which the evidence confirms a model have shown the need to rely
on extra assumptions which are themselves hard to justify \citep{confirmation-IEP}.  For these
reasons, predictions {\em alone}---no matter how impressive---are never sufficient to warrant
trust in a model, and Likelihood---by itself---is never sufficient to determine model selection.
Other non-empirical aspects are also important.  But which ones?

These problems also apply to TS models, but are more evident for DNN models whose persuasive power
relies relatively more on their empirical success than on their theoretical virtues.  Although DNN
successes are impressive, it is impossible to derive any quantitative level of confirmation from
them without additional assumptions.  For example, published results certainly suffer a positive
selection.  Kaggle competitions \citep{kaggle} are very interesting also from a philosophical
point of view because they provide a controlled framework for assessment.  However, they also
cannot be used directly to confirm either general DNN architectures or specific DNN models,
because submissions are strongly biased in terms of the approach used and the competencies of the
participants.  The fact that some successful DNN architectures keep being successful over time is
certainly convincing evidence that they can learn something valuable.  However, also in those
cases, we cannot exclude that the DNNs have actually learned unwanted features that happened to
have a strong spurious correlation with the labels (see, e.g., \citet{xiao2021noise} for an
example of this phenomenon).  This suggests that DNN predictions are only valuable when they stem
from a valuable model.  What characterises a valuable model, both in TS and in ML?  This question
is addressed in more depth in Sec.~\ref{sec:ass-int}.

\section{Assumptions, simplicity and interpretability}
\label{sec:ass-int}

The previous discussions started from different perspectives which all led to the same conclusion:
the reliability of a model necessarily depends on the reliability of all its model assumptions,
which are never self-justifying.  Empirical evidence alone is never enough, neither in ML nor in
TS.  In Sec.~\ref{sec:ass} we review which classes of assumptions characterise TS models, DNN
models and other ML models.  While we cannot assess the reliability of these assumptions, in
Sec.~\ref{sec:simp} we propose a measure of their simplicity as the most relevant surrogate used
in the scientific practice (often implicitly).  In Sec.~\ref{sec:interpret} we identify a close
relation between the simplicity of the assumptions and the concept of interpretability (now
intensively discussed within the field of responsible AI).

Any model, whether TS or ML is essentially characterised by all its assumptions and its accuracy
with respect to the existing empirical data\footnote{Defining a scientific or ML model by its
assumptions might suggest that our discussion is restricted to a {\em syntactic} approach to
scientific models.  However, following \citep{Lutz-ppr, Lutz-ss}, it also applies to a {\em
  semantic} approach.  Other views exist in the philosophical literature
\citep{sep-structure-scientific-theories}.  But, there is currently no compelling evidence of real
models that do not admit a formal description.}.  In the following, the expression ``all
assumptions'' refers to all those that are necessary to reproduce the outcome of the model,
including the comparison with the empirical evidence and the expression of the input
data\footnote{If the model predicts an outcome with some statistical distribution, as it is mostly
the case, the requirement is to reproduce the statistical distribution and not the individual
outcomes.}.  Normally, the training data themselves are not part of the model
assumptions\footnote{Note that the {\em input} data, instead, are part of the assumptions, but
they do not matter when comparing models on the same questions.  See Footnote \ref{note:input}.}:
they contribute to select the model parameters (which {\em are} part of the assumptions) and they
contribute to the empirical accuracy of the model just like the experiments that contribute to the
development of TS models.  However, including the training data among the assumptions might lead
to simpler formulations of some DDN models characterised by a huge number of parameters.  Hence,
this possibility is also considered.

\subsection{Assumptions}
\label{sec:ass}

TS models rely on multiple categories of assumptions: besides the specific assumptions of the {\em
  model} itself, they rely also on multiple scientific disciplines that are not the main focus of
the specific TS model but are essential to constrain the TS model.  These disciplines are referred
to as {\em background science} and they include a variety of domains from logic, mathematics,
basic physics, chemistry and the modelling of any relevant experimental device.  TS models
typically include also assumptions to restrict their own {\em domain} of
applicability\footnote{Note that scope restrictions can be expressed as part of the assumptions
(namely relativizations). Therefore, we will not discuss scope restrictions separately in this
paper.}.  In particular, TS models assume suitable values of the model
parameters\footnote{Alternatively, one might assume that the free parameters are determined by
Maximum Likelihood principle (in a frequentist approach) or via a Bayesian argument that includes
the assumption of Bayesian priors.  However, this requires also the assumption of the all the
empirical evidence necessary to compute the likelihood function, which is typically less concise
than assuming the values of the parameters (see Sec.~\ref{sec:simp} and Footnote
\ref{note:input}).}.  The parameters inherited from background science must be consistent with a
much broader spectrum of empirical evidence from very different domains.  In this sense,
traditional science adopts a kind of {\em divide et impera} strategy to parameter fitting.

On the other hand, DNN models require very little assumptions from background science (besides
modelling the data collection).  They are very generic and they are defined, ideally, only by the
architecture (hyper-)parameters and the domain in which they are trained.  However, this is not
precise enough to determine their predictions, not even statistically.  Predictions are certainly
determined by the entire set of DNN parameters, which are typically in the order of billions in
modern DNNs.  Ideally, the specific values of all those parameters are not essential to determine
a prediction, which should depend only on the training {\em domain}.  However, the training domain
is difficult to define and the outcome may depend also on subtle details of the training process.
In fact, weight initialisation and pre-training techniques are key design choices when training
and deploying a DNN model \citep{narkhede2022review, xavier-init}.  Adversarials
\citep{szegedy2014intriguing} are also evidence of high sensitivity to details.  If we had
evidence that the training of a DNN model depends only on some simple rules and is independent of
the specific initial values of the weights (within some clearly defined bound and approximations),
we would need to include only these rules as part of the assumptions of the DNN model.  Instead,
the lack of understanding of the training dynamics enforces the inclusion of the full
specification of the DNN (initial) parameters and training data as part of the assumptions.

It is interesting to compare the DNN training process to other scientific applications of Markov
Chain Monte Carlo (MCMC)\footnote{For other aspects of the comparison between DNN and computer
simulations, see also \citep{boge2022two}.}.  For many TS models, there is no {\em proof} of
convergence of the MCMC algorithm to a definite outcome.  However, best-practice rules and
diagnostic tools have been developed that enable a rather accurate formulation of the conditions
that must be fulfilled for the algorithm to converge to a stable outcome (see
e.g.~\citet{Roy2019MCMC-convergence}).  Formal proofs of independence from details, if available,
are very desirable, because no further assumption is needed in that case.  But, also assuming a
few semi-empirical rules is satisfactory, if they are de-facto accurate.  Despite considerable
effort, researchers have not been able to identify any simple set of rules that makes the outcome
of DNN models independent of any further detail.  Unfortunately, research is very difficult on
this topic, because it requires testing different initial conditions, which is computationally
very expensive.  Moreover, this difficulty might actually be an intrinsic price to pay for the
great flexibility of DNN models (see also \citet{hartmann2023transgressing}.  High sensitivity to
details is the key aspect of the so-called {\em black-box} problem \citep{Floridi2022-FLOTEF} of
some ML methods: {\bf lack of understanding is a serious limitation to the extent that we cannot
  tell which details actually matter for a conclusion}.

Furthermore, TS models strive to be consistent, when applied to different data and different
domains.  In particular, it is typically ensured that the values of the parameters with identical
meaning remain consistent, within expected uncertainties, across different applications.  Too
large deviations are interpreted as failures of the model.  DNN parameters, on the contrary, are
usually not regarded as something that should be consistent across applications.  Although DNN
training often starts from DNN models that were pre-trained on other datasets, further training is
always performed for new applications and no constraint is usually imposed to keep the parameters
close to the original values.  That means that different assumptions are made by the DNN models
for different domains of application.

A further key difference between TS and DNN models is the specification of the domain of
application.  TS models typically define their domain of applicability in terms of the features
that play a role in the model and they are typically measurable.  If the domain is defined in
terms of measurable features, it becomes possible to suspend a prediction for out-of-domain data.
For example, domain restrictions usually enter the detailed specifications of the experimental
set-ups that are required to collect valid data.  Normally, the domain of applicability can be
formulated in terms of a limited set of additional assumptions.  This does not ensure that some
overlooked features may not play an unexpected role and compromise the accuracy of the model, but
this scenario happens rarely for state-of-the-art TS models.

In the case of DNN models, a measurable specification of the domain of application is much more
difficult, and it is practically never provided.  One key idea of DNNs is that the relevant
features do not need to be specified explicitly.  But this complicates the formulation of the
domain of applicability.  At least the experimental set-up for valid data collection must be
specified with sufficient precision to ensure the relevance of the training dataset.  For DNN,
however, it is more difficult to define the domain by describing the experimental set-up, because,
again, we have a very limited understanding of which features are learned during the DNN training
process.  Hence, the risk of omitting relevant prescriptions is much higher than for typical TS
models.  If, on the other hand, we omit the specification of the domain, we should test the
performance of the DNN on any possible phenomena, even those totally different from the scope in
which the DNN was actually trained.

It is worth including LR models in this comparison because they represent an interesting
alternative to both TS and DNN models that live somewhere in the middle ground.  LR models are
defined unambiguously by their features\footnote{Normally, features are assumed to be directly
measurable for LR models.  If not, the assumptions of the LR model must include any specification
needed to define them in terms of measurable ones. See also App.~\ref{sec:complexity}.}.  Because
the optimal regression coefficients are uniquely defined by the Maximum Likelihood principle, no
coefficient must be included to define the assumptions of the model.  Domain restrictions are also
specified via the relevant features.

Another interesting comparison is with RF models: they also display significant sensitivity to
some details (initial conditions and algorithmic hyperparameters), but they are based on limited
and well-defined features.  This enables, in the first place, a clear definition of the domain of
applicability.  It also makes it easier to identify conditions of robustness, for some
applications and contexts.  A deep analysis of the stability of RF models is beyond the scope of
this article.  However, it should be noted that their sensitivity to details seems closer to
traditional applications of MCMC rather than DNN.

\subsection{Simplicity of the assumptions}
\label{sec:simp}

The previous section shows that a key difference between TS models and DNN models is that the
former use {\bf relatively few assumptions and few parameters for a wide range of phenomena}.  A
classic scientific-philosophical tradition singles out precisely this aspect as the main
non-empirical epistemic value that scientific models should try to achieve, in addition to
empirical accuracy (see e.g., \citet{GalileoDialogo}, p.~397, \citet{NewtonPrincipia}, p.~398,
\citet{Lavoisier}, p.~623, \citet{Poincare}, \citet{Mach}, \citet{weyl1932open},
\citet{EinsteinLife}, \citet{Kemeny1953TheUO}, \citet{QuineSTCW}, \citet{Lewis1973},
\citet{chaitin1975randomness}, \citet{Zenil2020demystify}, \citet{Walsh1979}, \citet{Derkse1992},
\citet{Swinburne1997}, \citet{NolanQP}, \citet{Scorzato}, also reviewed e.g. in \citet{Baker-SEP},
\citet{simplicity-IEP}, \citet{ZellnerKeuzenkampMcAleer2001}).  The first goal of this section is
to clarify the meaning of ``few assumptions''.

An apparent problem with this view is that the number (and the length) of the assumptions is
catastrophically language-dependent: it can always be made trivial---hence meaningless---by a
suitable choice of language\footnote{It is sometime noted that Kolmogorov complexity depends on
the language only via a constant \citep{LiVitanyi2019}.  However, this is not sufficient for the
purpose of model selection \citep{Votsis2016}.}.  However, it was observed in \citet{Scorzato}
that requiring the inclusion of a basis\footnote{A {\em basis} is a set of quantities assumed to
be measurable, that is sufficient to define all measurable quantities that are needed for the
comparison with existing empirical evidence.} of measurable concepts among the assumptions
enforces a minimal achievable complexity (see details in App.~\ref{sec:complexity} and
App.~\ref{sec:progress}).  The original argument of \citet{Scorzato} is applicable to a wide class
of TS models, but not immediately to DNN models.  It is shown in the App.~\ref{sec:complexity}
that the mere possibility of the existence of adversarials enables the extension of the argument
to DNN as well.  This enables the reference to the {\em epistemic complexity} of a model in a way
that is precise and language independent\footnote{Epistemic complexity depends, of course, on the
model and hence on its empirical content, which is not language dependent, but needs a language to
be described (see App.~\ref{sec:progress}).}.

How does the epistemic complexity of a model impact its reliability?  This question does not have
a simple answer.  Certainly, we cannot dismiss DNN models on the grounds of their high complexity,
because we mostly do not have simpler TS models that cover the same topics and complexity-based
model selection applies only to empirically equivalent models (see App.~\ref{sec:progress} for
more details).  However, a high complexity affects reliability at least in three indirect ways.
The first one is {\em interpretability}.  If it is challenging for any human to even understand
which are all the assumptions of the model or review them, it becomes difficult to even formulate
precisely the question: {\em what is it} whose reliability we are investigating?  In this sense, a
manageable epistemic complexity should be a precondition for a plausible assessment of
reliability.  The next Sec.~\ref{sec:interpret} looks in greater detail into the relation between
the epistemic complexity introduced here and the concept of interpretability that plays a
prominent role in the current discussion about responsible AI.

The second relation between simplicity and reliability comes from the fact that simplicity enables
the definition of the state-of-the-art models (see App.~\ref{sec:progress}), namely those models
that represent all the possible compromises between simplicity and the many dimensions of
accuracy.  The state-of-the-art, in turns, represents a non-arbitrary class of models that offers
a reference for a Bayesian framework for error estimation\footnote{In the scientific practice,
Bayesian estimates may not include the entire state-of-the-art, but the existence of the
state-of-the-art justifies other choices as approximations of a well defined class of models.}.
The high epistemic complexity of DNN models makes it very difficult to identify the
state-of-the-art.  The computational cost of the Bayesian approach applied to DNNs has been widely
acknowledged \citep{IEEE_errors_review, IF2021review, loftus2022uncertainty}, but it is important
to emphasise that this makes it also difficult to {\em define} the Bayesian errors in the first
place.

The third indirect relation between simplicity and reliability is via (scientific) {\em
  progress}.  It was observed already in \citep{Popper} that simpler models offer a more direct
path toward scientific progress.  The present framework offers further support for this idea:
simpler models offer fewer opportunities for small changes.  Hence, they point more clearly to
what must be changed, or they point to a fundamental deficiency of the model.  Complex models
might be more flexible and survive more empirical challenges, but that is not an advantage if we
are looking for a better model.  As a matter of fact, TS models are the result of a long history
of small and big unambiguous improvements, where different models are unified under a single one
(unification) or where the parameters of some empirical model are derived from a more fundamental
one (reduction) or radically new models supersede the the existing ones via revolutionary changes.
This history of progresses is certainly part of their perceived reliability.  On the other hand,
DNN models are too complex to allow the identification of this kind of unambiguous progress: every
fine tuning of paramters or architecture adjustment might bring some improvements, but it is
difficult to tell what might be lost.  This topic is discussed further in
Sec.~\ref{sec:conclusions}.

\subsection{Interpretability}
\label{sec:interpret}

Recently, the notions of interpretability, explainability, explicability, transparency and related
concepts have attracted considerable attention.  Together, they build the core taxonomy of the
topic of responsible AI \citep{arrieta2020explainable}.  There is still considerable debate and
obscurity around the precise definition of each of these terms, which have also started to acquire
a legal connotation.  The need for clarification has been analysed, from the philosophical point
of view, by \citet{beisbart2022interpretability}.  In fact, the concept of {\em interpretation}
has a long history in the philosophical literature (see, e.g., \citep{Lutz-int} as a starting
point).  It is not the goal of this paper to try to clarify the relations between these concepts.

Instead, we focus here only on what is relevant for reliability.  To this end, it is worth noting
that one of the most quoted definitions of {\em interpretability} reads ``the degree to which a
human can understand the cause of a decision'' \citep{miller2019explanation, lipton2018mythos}.
Here a ``decision'' translates to what we previously called a ``model prediction''.  What are the
causes of model predictions?  They are exactly the full set of model assumptions discussed in the
previous section.  In fact, they must include any information necessary to reproduce the model's
predictions\footnote{Another popular definition, that reads ``the ability to explain or to present
in understandable terms to a human'' \citep{doshi2017towards, hall2018art}, is more vague, but
essentially consistent with the one of \citep{miller2019explanation, lipton2018mythos}.}.
Anything less, would be an incomplete cause; anything more would be superfluous\footnote{The {\em
  interventionist} framework introduced by \citet{woodward2005making} is useful to identify the
causes of a {\em specific} output, which is relevant for {\em local} interpretations as in
\citep{watson2021explanation}.  However, when we consider the causes of {\em all} potential
outputs produced by a model, as we must do to assess its reliability, then all non superflous
model assumptions must be considered part of the causes.}.

It is essential to appreciate that what matters is to understand the {\em causes} (i.e.~the {\em
  model assumptions}) and not the entire {\em internal mechanism} that leads, step by step, from a
given input to the corresponding model output.  Most of our best scientific theories do not allow
such ``understanding'', which is often encoded in millions of operations executed by computers.
Not even the best world experts would be able to reproduce most of the model output without the
support of those computers.  But understanding these details is unnecessary\footnote{See also the
discussion in \citet{beisbart2021opacity} leading to similar conclusions concerning computer
simulations.}, because all these operations are entirely determined by a few equations, which are
the true {\em causes}, i.e.~the {\em model assumptions}\footnote{Knowledge of the model
assumptions is a precondition for counterfactual explanations \citep{baron2023explainable,
  buijsman2022defining}, but it is not sufficient to ensure it.  Full visibility of how the output
would change by any change of input may require unlimited computational power.  This holds for DNN
models as much as for most modern TS models.}.  Of course, this is true also because we know that
little details, such as the specific value of the random generator used in the simulation code, do
not matter.  Requiring a full understanding of the detailed mechanism is an act of desperation in
cases where we do not know which details actually matter, but it is not a sensible requirement
when we do know that.

The view described above is consistent with the one defended by \citet{deRegt2005-REGACA-2}, who
criticize the causal-mechanical conception of explanatory understanding for similar reasons to
those given here.  According to \citet{deRegt2005-REGACA-2}, experts are {\em often} able to
connect some changes in model assumptions to some changes in model outcomes, without performing
caluculations because they extrapolate from other results that they know.  But, for most modern
scientific models, even the best experts cannot predict most of the outcomes without the help of
external computational tool.

Another often-cited definition reads: ``providing either an understanding of the model mechanisms
and predictions, a visualisation of the model’s discrimination rules, or hints on what could
perturb the model'' \citep{arrieta2020explainable}.  As mentioned above, ``understanding the model
mechanisms'' (or visualising them) is too strong a requirement, that would unfairly penalize most
modern science.  However, this definition also emphasises an important aspect: control of ``what
can perturb the model''.  As discussed in the previous section, if a set of simple rules ensures
the stability of the output (not necessarily a unique exact output, but at least a unique
statistical distribution which does not depend on other details), then those rules are sufficient
causes/assumptions for the model's predictions.  If not, more details must be included among the
causes/assumptions.

Let us examine one more definition, which reads: ``a method is interpretable if a user can
correctly and efficiently predict the method’s results'' \citep{kim2016examples}.  This definition
tries to remove the ambiguity of the word ``understanding'', which is used elsewhere, but without
success: is the {\em user} allowed to utilise a tool to reproduce the model's results? If she is,
then she can just use another, identical, DNN model and every DNN model becomes perfectly
interpretable.  If she is not, then most modern science is not interpretable, according to this
definition, because no human being can go very far without any instrument.  This definition is not
ambiguous in the context of image recognition, where it has been introduced
\citep{kim2016examples}, because it assumes that human vision is the tool available to the user.
But, it cannot be extended to a general context, as suggested in \citep{molnar2020interpretable}.
However, this attempt contains a very interesting aspect: the concept of understanding must be
clarified; we must understand enough causes (i.e. model assumptions) to {\em reproduce} the
outcome, and the outcome must be ``consistent'', i.e. robust to changing details.  This is
achieved if we know all the model assumptions, if these do not require the inclusion of too many
details and if the output is computable for the relevant inputs (i.e. the input is legitimate for
the model).  So, also this last definition supports an identification of the concept of {\em
  interpretability} with a measure of the simplicity of all the assumptions.

In conclusion, we can attribute a precise meaning to the concept of ``interpretability''.  In this
way, ``interpretability'' does not depend on the vague notion of ``understanding'', does not
depend arbitrarily on the choice of language, and does not depend on the individual skills of the
persons interacting with the model.

Moreover, interpretability is not only a desirable property for non-expert, it is a precondition
for a plausible assessment of the reliability.  More precisely, lower interpretability makes the
assessment of reliability less plausible.  This claim can be justified as follows.  Assessing the
reliability of any model is done in two steps.  The first step consists in identifying a class
($C$) of models that should be used to assess the reliability of the given target model (see
Sec.~\ref{sec:BEE}).  The second step consists in computing the confidence ranges (e.g. via the
Bayesian priors and the Bayesian formula).  The first step is the focus of this paper.  In TS,
scientists tend to agree, de facto, on which class $C$ should be used (which we can call the
state-of-the-art models, for which \citet{Scorzato-silfs14} proposes a characterization.  See
Def.~\ref{def:progress} in App.~\ref{sec:progress}).  Whether we adopt Def.~\ref{def:progress} or
not, identifying the class $C$ requires examining non-empirical epistemic properties of the
corresponding models.  If we do adopt Def.~\ref{def:progress}, the state-of-the-art becomes
broader when more complex models are involved, because it is increasingly more difficult to check
if any of the many variations of a complex model should or should not be included in the class
$C$.  In fact, complex models leave larger opportunities for models that are not more complex.
This leads to larger confidence ranges and larger uncertainties on the confidence ranges
themselves.  Even if we did not adopt Def.~\ref{def:progress}, we should still evaluate some
(unclear) non-empirical properties of the models, which must include also reviewing their
assumptions.  The lack of precise requirements cannot simplify the assessment of more complex
assumptions.  This forces, again, larger uncertainties on the confidence ranges.  In both cases,
more complex (i.e.~less interpretable) assumptions lead to less plausible assessments of
reliability\footnote{Note that it is common practice to penalize complex assumptions by
reweighting the priors within a Bayesian framework \citep{Schwarz78}.  But it is also often
claimed that the effect of the priors vanishes for large statistics.  Here we discuss
uncertainties that are not eliminated statistically.}.

\section{Conclusions and way forward}
\label{sec:conclusions}

DNNs are being used effectively in many contexts where it is not necessary to assess their
reliability precisely.  This is the case when DNN models generate candidate solutions that are
eventually checked via independent tests \citep{AlphaFold}, or when they generate data that do not
need to be sampled with a strictly defined distribution, because the conclusion has no strict
statistical value or because the distribution is corrected afterwards \citep{Shanahan}\footnote{If
the independent check is performed by humans, it becomes important to compare the reliability of
DNN with the one of humans.  This a very important and interesting question, but beyond the scope
of this paper, that focuses on comparing DNN models to TS models.}.

However, when assessing the expected errors is necessary, it is important to understand how
traditional science achieves that.  The reliability of traditional science does not depend only on
a statistical analysis of the uncertainties.  It depends also on the fact that scientific models
rely on {\em few assumptions} that remain the same for a very large amount of phenomena.  This is
accomplished by building minimal models for different domains of phenomena.  Of course, these
different domains are interdependent and they must be combined to describe more complex phenomena.
However, this {\em divide et impera} strategy turns out to be quite efficient.  Moreover, TS
models strive to employ as much analytical understanding as possible (enabled by simpler models),
next to empirical testing to ensure that empirical successes aren't ephemeral.  In other words,
{\bf this strategy enables progress towards models that are gradually more and more accurate
  and/or more and more interpretable (i.e. simpler)}.

DNNs often display impressive predictive power, which can only be possible because, somehow, the
training process finds a DNN configuration that is sensitive to the relevant features of the data
and is insensitive to the irrelevant ones\footnote{The observation that DNN seem to be able to
identify the right features, sometimes, is equivalent to state that it seems to solve the {\em
  reference class problem}, sometimes \citep{buchholz2023deep}.  However, our current problem is
to characterise {\em sometimes}.}.  However, while the {\em existence} of such configurations (and
the ability of the algorithm to {\em find} them) is sufficient for their powerful predictivity, it
is not sufficient for their reliability.  To define what we mean by reliability, we must clarify
the class of possible alternative models (see Sec.~\ref{sec:BEE}).  TS models can typically refer
to the class of models that belong to the state-of-the-art or some approximations of it.
Identifying the state-of-the-art is much more difficult for DNN models because of their high
epistemic complexity, which makes it difficult not only to compute the Bayesian errors, but also
to define them.

This paper has proposed to identify the interpretability of a model with its inverse epistemic
complexity.  In this sense, interpretability is not only a desirable property for non-expert
users: it is a precondition for a plausible assessment of the reliability itself.  Identifying the
assumptions is also necessary for scientific progress \citep{Scorzato-silfs14}, because we need to
know which assumptions we may want to replace.

Impressive predictive power in new domains, combined with weak foundations and difficult synthesis
with background-science can be a signal of an ongoing (and incomplete) scientific revolution, as
was the case for the birth of Quantum Mechanics.  What could be the way forward?  Predicting the
retirement of the scientific method itself (towards something like a theory-free science) appears
misguided, according to the present analysis.

One natural hope is to better understand the conditions that ensure the robustness of DNN
predictions, hence enabling an interpretable formulation of the assumptions.  However, this might
be inherently impossible, because the high flexibility of the DNN might be intrinsically coupled
with their high sensitivity to details \citep{hartmann2023transgressing}.

One interesting alternative approach is to use DNN as a tool to suggest features for other models,
along the lines of \citep{huang2006large, notley2018examining}.  However, it is important to
ensure that the features obtained through this process are directly measurable and not simply the
output of the last DNN layer, otherwise the DNN component cannot be removed from the formal
counting of the assumptions needed to produce the results.  The features extracted in this way
would ease the connection with background science (imposing valuable constraints) and enable the
building of simpler models, possibly unrelated to DNN models.
See \citep{guth2023rainbow} and references therein for significant progresses in this direction.

Another interesting approach is to focus on those models where the entire possible input space is
potentially completely available for training.  This is the case, for instance, when reading
printed characters (from a limited choice of fonts), or scanning objects that belong to a finite
list of possibilities (possibly rotated and translated in space).  To cover the full input space,
it is crucial to identify and implement all possible symmetry transformations, which include not
only spatial transformations but also background and noise transformations.  Even when all these
steps are put in place, it is still difficult to check that the entire possible input space has
been covered.  Adversarials may still lurk, as long as the DNN is not exactly invariant under the
transformation of redundant parameters.  Ideally, to ensure robustness, one should also reduce the
number of DNN parameters to the actual degrees of freedom of the input space.  This is not easy,
but it might be possible along the lines of \citep{Liu_2015_CVPR}.

Another promising avenue of research is to study those limits that can be computed exactly, as the
limit of infinitely large width of fully connected DNNs \citep{Jacot2018NTK}.  Such analytical
results are essential to test the behaviour of a DNN where we know exactly how it should behave.
So, their first epistemological advantage is to increase the empirical accuracy of the DNNs (or
rule them out).  Furthermore, these limit cases could potentially also suggest how to realise
simpler ML methods that are as powerful as DNN under specific circumstances.  Another possibility
is to use exactly solved limit cases as a starting point to set up DNNs that are small deviations
from those limit cases, which might potentially enable a simpler formulation (with fewer
assumptions) of the DNN itself.

\appendix
\section{On the irreducible complexity of DNN representations}
\label{sec:complexity}

One may question the meaningfulness of assessing the complexity of the assumptions by counting the
words used to express them.  In fact, the length of the expression of a set of axioms is
language-dependent.  Even worse, for any set of axioms, one can always find a language where the
same axioms assume a trivially concise form \citep{Kelly-razor, Votsis2016}.  This is the origin
of an old paradox: why do scientists employ relatively complex formulations, even for their most
fundamental theories, when a much simpler one is available?  Is it due to hidden cultural bias?
An answer was given in \citet{Scorzato}, where it was shown that the language that makes the
axioms trivially short typically employs concepts that are not {\em measurable}.  If we require
the axioms to include sufficient relations to measurable concepts, finding a concise formulation
becomes non-trivial and the shortest formulation among all possible languages becomes meaningful.
In fact, the standard scientific formulations are typically the most concise that we can find,
under these constraints\footnote{Indeed, the scientific language evolves with science in a global
optimisation process \citep{Kvasz-Pattterns}.}.

The argument in \citet{Scorzato} applies only to scientific theories that are sufficiently complex
to allow the emergence of chaotic phenomena.  In particular, any scientific theory that includes
even the simplest form of Newton's equations (or similar differential equations) is already
complex enough to fall within the scope of the argument in \citet{Scorzato}.  Because most TS
models typically rely on background science and the latter typically includes at least some
minimal notions of basic physics, it follows that, for most real-life TS models, the length of the
assumptions in their native scientific language is a meaningful measure of complexity.

However, DNN models rely on very little or no background science.  Hence, it is an interesting
question whether the formulation of a DNN model can be made arbitrarily concise or not.
Interestingly, the mere possibility of the existence of adversarial examples enables an argument
similar to the one in \citet{Scorzato}.  This is not surprising, because adversarial examples have
strong analogies with chaotic phenomena.  As we will show below, a coordinate system does exist
that makes the formulation of the DNN trivially concise (with just one parameter) and it is easy
to find.  But that formulation is not measurable (in the sense specified below).  Moreover,
finding a coordinate system that makes the DNN significantly more concise than the standard
formulation, while provably preserving measurability, is a major challenge with no known solution.
Therefore, the best estimate we have of the complexity of a DNN is the one that we can derive from
its standard formulation.

Let us see how a simple formulation may run into conflict with measurability.  It is not difficult
to find a coordinate system $\xi \in [0,1]^n$ describing the objects to be classified (e.g. images
of dogs and cats) such that the DNN classifies an image as cat (dog) when $\xi_0=0$ ($\xi_0=1$)
and classifies it with decreasing (increasing) probability as cat (dog) when $\xi_0$ increases
from $0$ to $1$.  We can not only prove that such coordinates exist: they are essentially
equivalent (up to trivial transformations) to the last layer representation of a DNN.  In these
coordinates, the DNN would have an extremely simple representation.  In fact, it would have just
one parameter $w=1/2$ and the final output would be {\tt cat} if $\xi_0<w$ and {\tt dog}
otherwise.

Note that the $\xi_0$ coordinate does not tell whether the image represents a cat or a dog, it
tells whether the DNN {\em predicts} it to be a cat or a dog (we will call these images {\em
  p-cats} and {\em p-dogs} for short).  In fact, the more concise formulation that employs $\xi$
does not change the predictivity of the DNN model, it only re-expresses it with a simpler
notation.

Now, are the concepts of p-cat-ness and p-dog-ness {\em measurable}?  In a model for visual
perception of images, any measurement must be based on the observation of the colour of each
picture element.  The colour and position of each picture element can be determined only up to
some limited precision, that is defined by human sensitivity.  Hence, the claim that the
attributes of p-cat-ness and p-dog-ness are measurable may be plausible only if small {\em
  imperceptible} changes in the picture cannot induce a change on their assessment from almost
certainly p-cat ($\xi_0\simeq 0$) to almost certainly p-dog ($\xi_0\simeq 1$).  However, this is
exactly what the adversarials do: they are instances very close to p-cats which are 
humanly indistinguishable from dogs, and actually very close to p-dogs.  In this sense, the
concepts of p-cat-ness and p-dog-ness cannot be measurable.  Here we have assumed that {\em
  measurable} quantities cannot assume different values, with high confidence, for imperceptibly
different data points.

Note that we have {\em not} proved that any DNN model {\em cannot} be represented with coordinates
that are both very concise and also measurable.  We have only shown that an easy, very concise,
formulation is not measurable.  But this is sufficient for our goals, namely to refute the claim
that there is always a way to formally trivialise the formulations of any system of axioms
\citep{Kelly-razor, Votsis2016}.  Although that claim remains correct from a logical point of
view, it does not hold anymore once the constraints of measurability are enforced.  Discovering a
simpler formulation might be possible and it would be a significant achievement\footnote{Note also
that, in general, it is not possible to tell whether a given formulation is the shortest possible:
this question is not computable, as shown in \citet{Chaitin}.  However, that simply means that the
complexity of a real scientific model can be only {\em estimated approximatively}, as it is the
case for most scientific quantities.}.

\section{A simple model of progress}
\label{sec:progress}

This appendix specifies the concepts of {\em (epistemic) complexity}, {\em theory selection} and
{\em progress} introduced in the main text and it is essentially a summary of
\citep{Scorzato-silfs14}.  The first step is to define a rule for {\em model selection}.  We have
seen that {\em empirical accuracy} is not enough to determine model selection: there are always
infinite empirically equivalent theories.  The key is then to identify at least one\footnote{One
single additional value, besides empirical accuracy, offers an acceptable simple model of
progress.  See \citep{Scorzato} for a discussion of other potential epistemic values that appear
to be non-independent from simplicity and empirical accuracy.}  non-empirical epistemic value that
is sufficient to rule out empirically equivalent alternatives that the scientists would not
consider valuable\footnote{This criterion is {\em descriptive}.  But, like any scientific law, it
might be considered {\em normative} to the extent that it represents a good description.}.

The idea that {\em simpler assumptions} should be preferred to more complex ones---if they are
empirically equivalent---is supported by a strong scientific/philosophical school of thought (see
e.g., \citet{GalileoDialogo}, p.~397, \citet{NewtonPrincipia}, p.~398, \citet{Lavoisier}, p.~623,
\citet{Poincare}, \citet{Mach}, \citet{EinsteinLife}, \citet{weyl1932open}, \citet{QuineSTCW},
\citet{Kemeny1953TheUO}, \citet{Lewis1973}, \citet{Walsh1979}, \citet{Derkse1992},
\citet{Swinburne1997}, \citet{NolanQP}, also reviewed e.g. in \citet{Baker-SEP},
\citet{simplicity-IEP}, \citet{ZellnerKeuzenkampMcAleer2001}).

A natural way to quantify the simplicity of the assumptions of a model is via Kolmogorov
complexity (KC) \citep{Kolmogorov, Chaitin, Solomonoff, Zenil2020review, Hutter04uai, Votsis2016,
  LiVitanyi2019}.  In general, the KC of a data string $D$ is defined as the length of the
shortest program $p$ (in a given programming language) that produces $D$.  In this context, the KC
simply measures the length of the assumptions of the model ${\cal M}$ in the shortest formulation
available in the given formal language \citep{chaitin1975randomness, Scorzato}.  Comments about
other frameworks can be found in the references cited in the App.~\ref{sec:beyond}.

A key question is the dependence on the language in which ${\cal M}$ is formulated.  In
App.~\ref{sec:complexity} it is shown that, if we require that ${\cal M}$ includes a definition of
a basis\footnote{A {\em basis} is a set of quantities assumed to be measurable, that is sufficient
to define all measurable quantities that are needed for the comparison with existing empirical
evidence.} of measurable quantities, there is no practical way to find a language that makes the
Kolmogorov complexity of ${\cal M}$ trivial.  This extends to DNN models a similar argument for TS
models \citep{Scorzato} and justifies the definition of the complexity of ${\cal M}$ as the
minimum length of the axioms across all available (measurable) formulations.  See
App.~\ref{sec:complexity} for more details.  Accordingly, we define:
\begin{defn}
  \label{def:complexity}
  The {\bf (epistemic) complexity} of a model ${\cal M}$ is the minimal length---across all
  possible formulations (in any language) of ${\cal M}$---of all the assumptions of ${\cal M}$
  that are needed to derive all the available results of ${\cal M}$.  The assumptions must contain
  all references needed to ensure that any comparison between the empirical data $D$ and the
  corresponding results of ${\cal M}$ are measurable\footnote{For a discussion of the concept of
  ``measurable'' see App.~\ref{sec:complexity} and \citep{Scorzato}.}.  The {\bf (epistemic)
    simplicity} of a model ${\cal M}$ is the inverse of its complexity.
\end{defn}
Being defined as the minimum across all possible formulations, Def.~\ref{def:complexity} is
obviously language independent.  It is a property of the model and it therefore depends on the
concepts that define the empirical content of the model, but it does not have any catastrophic
dependence on the language.  In fact, the interesting observation is that it is also in
non-trivial, in general, thanks to the analysis in App.~\ref{sec:complexity}.

The same analysis justifies an estimate of the complexity by relying on the natural language in
which the assumptions of a model are originally formulated \citep{Kvasz-Pattterns}.  In the same
spirit, we can assume a standard alphabet of symbols that can be used to express rather
efficiently any existing model.  A rough estimate of the model complexity can then be done by
estimating the number of characters that are necessary to express the full assumptions.  To be
concrete, the specific model assumptions of a typical TS model might include about $O(10-1k)$, to
which we must add, say, $O(10)$ scientific models from background science\footnote{The advantage
of using background science is that the same assumptions cover a much wider scope of phenomena.},
which leads to a total of $O(100-10k)$ characters.

Model selection can then be defined as a Pareto efficient selection of models, where the Pareto
agents are represented by simplicity and all the dimensions of empirical accuracy.  Specifically,
\begin{defn}
  \label{def:selection}
{\bf (Model selection)} Given a set of empirical questions (i.e.~a topic $t$), a model $A$ is
preferred to model $B$ if $A$ is neither more complex nor less empirically accurate than $B$ on
the topic $t$, while being strictly better than $B$ in at least one of these aspects.  In this
case, we say also that model $A$ is {\bf better} than model $B$ and $B$ is {\bf worse} than
$A$\footnote{Note that, when comparing two models on the same data, the expression of the input
data should be included as part of the assumptions, to prevent the possibility that one model
artificially pushes the model complexity in the expression of the input data.  This technical
remark is only needed to ensure that the complexity of the input data does not actually
matter. \label{note:input}}.
\end{defn}
Note that this model selection does not require any trade-off\footnote{Well known rules exist that
do define trade-offs between accuracy and model size (applicable for similar models that differ
only in their dimension) \citep{AIC, Schwarz78, CIC}.  Different rules correspond to different
trade-offs defined by different choices of priors \citep{BurnhamAnderson2004AIC_BIC}.  The
epistemic complexity defined above extends the measure of complexity to any model, but it refrains
from proposing a trade-off, which necessarily contain an additional arbitrary choice.}: it defines
a Pareto frontier (state-of-the-art)\footnote{Note that for two models entailing different
empirical consequences, their comparison in terms of simplicity does not lead to a selection.
However, it is often possible to alter ad-hoc the assumptions on both models to make them both
empirically accurate and equivalent.  The new (more complex) models are then comparable in terms
of simplicity and the comparison may lead to model selection if one is better than the others
beyond approximation errors.}.  Progress occurs whenever a Pareto improvement is achieved, thanks
to a new model or new measurements:
\begin{defn}
  \label{def:progress}
  Given a topic $t$, the {\bf State-of-the-art} is the ensemble of models which are not worse than
  any other model for the topic $t$.  The models that are not state-of-the-art are {\bf obsolete}.
  There is {\bf progress} when a model that was in the state-of-the-art becomes enduringly
  obsolete.
\end{defn}
These definitions are compared to a variety of real cases of theory selections and scientific
progress in \citet{Scorzato-silfs14}.

\section{Alternative notions of complexity}
\label{sec:beyond}

Kolmogorov complexity is a very popular measure of complexity, also widely used to estimate the
complexity of a theory \citep{chaitin1975randomness}.  The analysis in \citep{Zenil2017smalldata}
goes into great details to justify why Kolmogorov complexity is well suited to estimate the
complexity of both scientific and any ML model.  Nevertheless, it is not the only possibility and
one may question how the present analysis would change if we had used a different measure.  For
example, any specific compression software, like, e.g., gzip, implicitly defines an alternative
notion of complexity.  However, in all these cases, our conclusions about the complexity of DNN
models would not change\footnote{See, however the discussion in \citep{Zenil2020demystify} that
emphasises how compression techniques miss important features that are captured by Kolmogorov
complexity.}.  Counting the atomic sentences \citep{rus-met} would also not change our
conclusions.

In a ML context, it is common to refer to the {\em VC dimension} \citep{VCdim1971,
  harman2012reliable}, which is also interpreted as a measure of the complexity of the hypothesis
\citep{vapnik1999nature, corfield2009falsificationism}.  To the extent that the VC dimension
actually measures the complexity of the hypothesis, it is equivalent to our
Def.~\ref{def:complexity}.  However, such interpretation has major limitations, if applied to our
context.  In this context we need a measure of {\em all} the assumptions of {\em any} model, in
the language that is optimal for the model itself.  The VC dimension is designed for ML models
only.  In particular, the VC dimension of any parameter-free model is zero, which is very
misleading for TS models that contain many ad-hoc hypothesis.  Even if restricted to ML models,
the VC dimension has problems of both overestimation and underestimation of the real complexity of
the model assumptions.  Concerning the former, it has been noted that the VC dimension is not
suitable to describe the complexity of DNN models that contain a very large number of parameters
\citep{dziugaite2017computing}.  This is the case whenever the algorithm is able to reduce the
possible configuration that the model is able to reduce in practice.  On the other hand, the VC
dimension does not cover all those hypothesis that are not expressed in terms of explit parameters
in the ML model.  A number of extensions have been proposed \citep{barlett1998,
  bartlett2002rademacher, wang2022confidence} and, as long as they cover {\em all} the model
assumptions, they are implicitly included among the possible expressions over which the minimum of
Def.~\ref{def:complexity} is computed.  A similar comment is applicable to the model dimension
that is used by classic rules that define a trade-off between accuracy and complexity \citep{AIC,
  Schwarz78}.

One may also question the choice of using {\em any} notion of complexity of the assumptions to
select among empirically equivalent models.  In fact, philosophers of science have proposed a wide
range of alternative epistemic values potentially relevant for theory selection.  However,
\citet{Scorzato} argues that also these different values---to be meaningful---implicitly rely on
the existence of a non-trivial notion of the complexity of the assumptions.  Moreover, once the
constraints of measurability are imposed, most classic epistemic values turn out to be not
independent of epistemic complexity and empirical accuracy.  The question is not settled, but we
take the current evidence to justify the restriction of this paper.

\paragraph{Disclosure and Disclaimer}
The author works for a company that undertakes business in the deployment of AI systems as part of
its commercial activities. The views expressed in this article are those of the author alone and
do not necessarily represent the views of his employer.

\bibliography{philo, ML}{}
\bibliographystyle{chicago}

\end{document}